\documentclass[journal]{IEEEtranTIE}

\usepackage{graphicx}
\usepackage{cite}
\usepackage{picinpar}
\usepackage{amsthm,amsmath,amssymb}
\usepackage{url}
\usepackage{flushend}
\usepackage[latin1]{inputenc}
\usepackage{colortbl}
\usepackage{soul}
\usepackage{multirow}
\usepackage{pifont}
\usepackage{color}
\usepackage{alltt}
\usepackage[hidelinks]{hyperref}
\usepackage{enumerate}
\usepackage{siunitx}
\usepackage{breakurl}
\usepackage{epstopdf}
\usepackage{pbox}

\usepackage{mathrsfs}
\usepackage{mathtools}

\usepackage{xcolor}
\usepackage{xpatch}

\usepackage[ruled,linesnumbered]{algorithm2e}

\graphicspath{{figures/}}

\newtheorem{lemma}{Lemma}

\newtheorem{theorem}{Theorem}

\newtheorem{assumption}{Assumption}

\begin{document}

\title{Distributed Formation Shape Control of \\ Identity-less Robot Swarms}

\author{
	\vskip 1em	
	Guibin~Sun, 
	Yang~Xu, 
	Kexin~Liu, 
	and Jinhu L\"{u}, \emph{Fellow, IEEE}

	\thanks{	
		This work was supported in part by the National Key R\&D Program of China under Grant 2022YFB3305600, in part by the National Natural Science Foundation of China under Grants 92067204 and 62141604, in part by the National Key Laboratory of Multi-perch Vehicle Driving Systems under Grant QDXT-NZ-202407-01, in part by the China Postdoctoral Science Foundation under Grant 2023M740185, in part by the Beijing Natural Science Foundation under Grant QY24135, and in part by the Beihang Zhuoyue Postdoctoral Fellow Program. 
		\emph{(Corresponding author: Kexin~Liu.)}
		
		Guibin Sun and Yang Xu are with the School of Automation Science and Electrical Engineering, Beihang University, Beijing, China (e-mail: sunguibinx@buaa.edu.cn, xuyangx@buaa.edu.cn). 

		Kexin~Liu and Jinhu L\"{u} are with the School of Automation Science and Electrical Engineering, Beihang University, Beijing, China, and also with the Zhongguancun Laboratory, Beijing, China (e-mail: kxliu@buaa.edu.cn, jhlu@iss.ac.cn).
	}
}

\maketitle

\begin{abstract}
	Different from most of the formation strategies where robots require unique labels to identify topological neighbors to satisfy the predefined shape constraints, we here study the problem of identity-less distributed shape formation in homogeneous swarms, which is rarely studied in the literature. 
	The absence of identities creates a unique challenge: how to design appropriate target formations and local behaviors that are suitable for identity-less formation shape control.  
	To address this challenge, we propose the following novel results. 	
	First, to avoid using unique identities, we propose a dynamic formation description method and solve the formation consensus of robots in a locally distributed manner. 
	Second, to handle identity-less distributed formations, we propose a fully distributed control law for homogeneous swarms based on locally sensed information. 
	While the existing methods are applicable to simple cases where the target formation is stationary, ours can tackle more general maneuvering formations such as translation, rotation, or even shape deformation. 
	Both numerical simulation and flight experiment are presented to verify the effectiveness and robustness of our proposed formation strategy. 
\end{abstract}



\definecolor{limegreen}{rgb}{0.2, 0.8, 0.2}
\definecolor{forestgreen}{rgb}{0.13, 0.55, 0.13}
\definecolor{greenhtml}{rgb}{0.0, 0.5, 0.0}

\section{Introduction}

\IEEEPARstart{D}{istributed} formation control of robot swarms has been studied from various aspects in the past two decades due to its important applications in many fields \cite{Harikumar2019Firefight,Lopez2020Enclose,Dai2023formation}. 
The objective of formation control is to drive a group of robots starting from any given initial configuration to form a target geometric shape. 
To achieve this objective, various control strategies have been explored, such as graph-based \cite{Zhao2019Bearing,Ze2024TIE} and assignment based \cite{Morgan2016Assignment,Sakurama2020Assignment} strategies. 
However, these state-of-the-arts still face fundamental limitations in terms of fault tolerance and group scale. 
It is related to a key attribute of the robots: Identity, which characterizes whether the individuals of a group are functionally interchangeable with each other. 

The employment of unique identities for robots in a group can simplify the design of the collaborative strategy for formation control. 
For instance, graph-based formation is simply achieved by controlling the feedback of the errors between shape constraints and inter-robot relative states in the presence of unique identities \cite{Sakurama2020Assignment}. 
The shape constraints are predefined by inter-robot relative position \cite{Nguyen2020Localization}, relative distance \cite{Bae2021Distance}, and relative bearing \cite{Zhao2019Bearing}. 
Since the constraints are usually described by a graph, such a strategy is so-called graph-based formation. 
This control strategy requires each robot to possess a unique identity and recognize its neighbors by the identities. 
Once robots know the identities of their neighbors, the consequent task is simply to obtain the desired constraints and current states relative to their neighbors and feedback on the errors between these two quantities \cite{Sakurama2020Assignment}. 
Assignment-based formation is another strategy that requires the use of unique identities for robots. 
This strategy adopts a procedure of assignment in either centralized \cite{Alonso2012Animation,Yu2013CDC} or distributed manners \cite{Morgan2016Assignment,Sakurama2021Coordination}. 
In the process of assignment, robots are required to intermittently or continuously identify and match between themselves and goal locations by using unique identities such as globally unique identities \cite{Alonso2012Animation,Morgan2016Assignment} or locally unique identities \cite{Sakurama2020Assignment,Sakurama2021Coordination}. 
The local identity is used more flexibly than the global one, however, it requires robots to use complicated rules to label their neighbors in a locally maximal clique (i.e., complete sub-graph). 
Once robots are assigned unique goal locations in a target formation, the subsequent task is simply to plan collision-free paths for them to reach their corresponding goal locations \cite{Wang2020Formation}, which greatly reduces the task complexity. 

The requirement of unique identities for robots may bring a series of problems as follows. 
First, due to their unique identities, the robots in a group are not interchangeable with each other.  
As a result, if a robot breaks and has to leave the formation, other robots are not able to replace the broken one seamlessly even though all the robots are the same in terms of hardware and functions. 
At this time, additional algorithms such as task reallocation or formation redesign are needed to deal with this situation \cite{Zhao2019Rigidity,Li2022Bearing,Chen2023Formation}. 
In addition, graph-based formation still needs to address the dimensionality issue due to the large-size shape graph of large-scale swarms \cite{Zheng2022TAC}. 
To sum up, this requirement lacks adaptability and robustness against group-scale variants. 
Second, the robot may not be able to obtain the identities of its neighbors. 
In practice, the most common way to obtain the robot identities is through a wireless network. 
For example, each robot can use its IP address as a unique identity. 
However, when wireless communication is not available and each robot is only able to use onboard sensing such as vision to detect its neighbors, such identities are difficult to obtain because it is challenging for vision to distinguish massive robots with similar appearances \cite{Sakurama2020Assignment}. 

Motivated by the above analysis, we focus on the problem of distributed formation shape control without robot identities. 
This problem has been addressed in very few existing studies. 
One representative work that can solve this problem is the edge-following strategy \cite{Rubenstein2014Assembly}, whose motion efficiency is however low because it only allows the robots on the edge of a swarm to move while all the others must stay stationary \cite{Sun2023Aseembly}. 
Another identity-less work employs a phototaxis-inspired mechanism for massive shape formation based on the concept of artificial light field \cite{Chu2023TASE}. 
To calculate the light field, each robot needs to acquire the positions of all the other robots, which is difficult in practice because robots usually have limited perception capabilities. 
In contrast to \cite{Chu2023TASE}, the work in \cite{Wang2020Formation} proposes a fully distributed strategy to realize shape formations by using local task swapping. 
However, this work also faces the problem of low motion efficiency because robots must adopt grid-ifying motion to provide both conflict-free and collision-free guarantees. 

Despite the existing studies, the following challenges remain open to overcome. 
The first important problem is how to describe the target formation to avoid the use of unique identities. 
The most intuitive approach is by potential fields \cite{Gazi2005Potential,Hsieh2008Potential,Bi2018Formation}. 
One limitation of this class of approaches is that robots may easily get trapped in local minima, making it difficult to form complex shapes \cite{Wang2020Formation}. 
Another class of approaches are to discretize the target shape into a sequence of goal locations and allocate the goals to each robot through local assignments. 
However, since matching and assignment problems need to be solved, identities are inevitably introduced.  
The second practical problem is how to design distributed control law to achieve identity-less formations. 
An awesome choice is based on the locally defined interaction behaviors. 
The behavior-based method has been widely studied since Reynolds rules were proposed in \cite{Reynolds1987ASCG}. 
In such a method, the target shape is described by an attractive potential field \cite{Hou2012ICTA,Vickery2021ICARA}. 
Each robot enters the shape by the attractive behavior generated by the shape potential, and avoids collisions with other robots through repulsive behavior.
However, The attraction-repulsion method may easily get stuck in local minima, making it difficult to form complex shapes \cite{Sun2023Aseembly}. 

The contribution and novelty of this paper are summarized below. 
First, we introduce a graphical shape description method that can be used for identity-less distributed formation, and propose a distributed negotiation mechanism that allows all the robots to reach a consensus on the target formation based on their neighbors' interpretations of the formation. 
This negotiation mechanism can well balance the local interest and constraints of different robots. 
Second, we propose a distributed behavior-based control law for identity-less shape formations of second-order robot swarms. 
With the proposed strategy, a robot swarm can form nonconvex complex shapes with high motion efficiency and adaptability to group-scale variants, as verified by comparison and experiment results. 
In addition to forming the target shape, formation maneuver control, such as translation, rotation, or even shape deformation of the formation, can be tackled. 
Being able to track maneuvering formations for massive robots is a vital feature of our proposed strategy. 
In contrast, the state-of-the-art methods for massive shape formation are only applicable to stationary ones \cite{Rubenstein2014Assembly,Wang2020Formation,Chu2023TASE}. 
Finally, both numerical simulation and experiment results are presented to verify the effectiveness and robustness of our proposed strategy. 
A representative experiment by changing the group scale demonstrates the high robustness of our identity-less strategy against newly added robots and robots' failures. 

The remainder of the paper is organized as follows. 
Section \ref{Sec_statement} introduces the problem formulation. 
Section \ref{Sec_formation} presents the dynamic formation description method. 
Section \ref{Sec_controller} addresses identity-less distributed formation shape control of homogeneous swarms. 
Simulation and experiment results are given in Section \ref{Sec_results}. 
Conclusions are drawn in Section \ref{Sec_conclusion}. 

\begin{figure*}[!t]
	\centering
	\includegraphics[width=17.9cm]{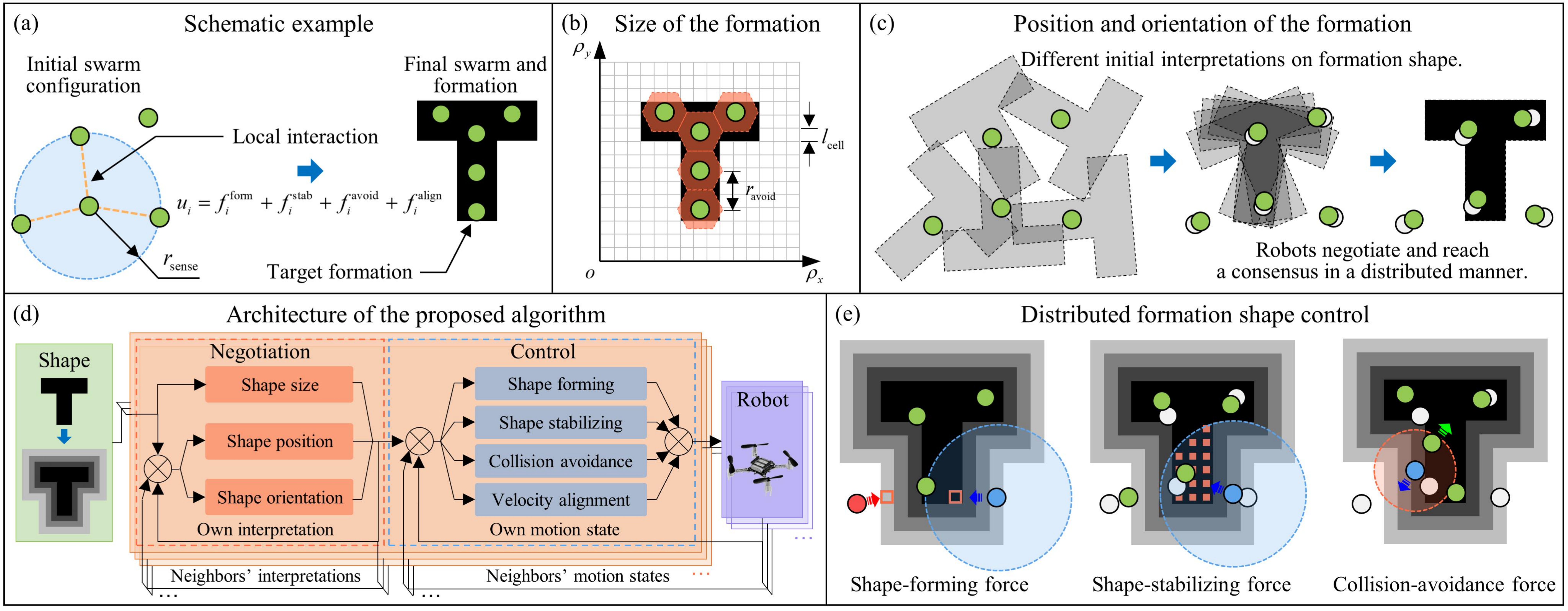}
	\caption{An overview of the proposed control strategy. 
	(a)~Schematic example of formation shape control. 
	(b)~Graphical target formation shape specified by users. 
	(c)~Example to illustrate the formation negotiation process of robots. 
	Each robot initially has different interpretations on the shape, and eventually reaches a consensus with the negotiation protocols. 
	(d)~Multiprocess implementation structure. Every robot corresponds to a unique running process. 
	(e)~Example to illustrate the formation control process of robots. Details of control command can be found in Section~\ref{Subsec_formcontrol}. }
	\label{Fig_overview}
\end{figure*}

\section{Problem Statement}
\label{Sec_statement}

Consider $n_{\rm robot}$ mobile robots in $\mathbb{R}^d$ and $n_{\rm robot} \ge 2$. 
Each robot is regarded as a sphere with a radius as $r_{\rm body}$.
Let $p_i\in \mathbb{R}^d$ be the position of the center point of robot $i$ in a global coordinate frame. The dynamic model is $\ddot p_i = u_i, i=1,\dots,n$, where $u_i$ is the control command to be designed. 
This paper mainly considers the case of $d=2$. 
When the distance between two robots is less than a threshold $r_{\rm sense}$, the two robots could share information with each other via passive sensing or wireless communication. The information network defines an undirected graph $\mathcal{G} = (\mathcal{V},\mathcal{E})$, which consists of a vertex set~$\mathcal{V} = \{1,...,n_{\rm robot}\}$ and an edge set $\mathcal{E} \subseteq \mathcal{V} \times \mathcal{V}$ such that~$\mathcal{E} = \{(i,j): \|p_i-p_j\| < r_{\rm sense},  j \neq i\}$. Here, $\|\cdot\|$ is the Euclidean norm. The set of neighbors of robot~$i$ is $\mathcal{N }_i=\{j \in \mathcal{V}: (i,j) \in \mathcal{E}\}$.

The control objective is to steer a group of robots starting from an initial configuration to form a user-specified formation shape through local interactions (see Fig.~\ref{Fig_overview}(a)). In the target formation, there is no one-to-one correspondence between robots and their target positions. 
The robots should be distributed as evenly as possible in the formation. 
They are not required to form any precise relative positions as long as the overall formation shape is consistent with the expected. 
In a sense, our research focuses on the full cover of the target shape by robots, regardless of the robot's specific position in the shape. In this way, the robot swarm can exhibit strong robustness to group-scale variants due to there being no correspondence between robots and the target shape, as verified by the experiments. 
Moreover, the entire group should be able to move collectively following a time-varying reference while maintaining the target formation shape. 

One may notice that the control objective or the target formation is not precisely defined compared to other formation problems \cite{Zhao2018Affine,Wang2020Formation}. 
That is mainly due to the lack of correspondence between robots and the target formation. 
It is worth emphasizing that the precise formation like \cite{Zhao2018Affine,Wang2020Formation} is not within the scope of this paper. 
Next, we show how to design an appropriate target formation to achieve such an objective. 

\section{Target Formation and Parameterization}
\label{Sec_formation}

This section addresses how to design appropriate target formations that are suitable for identity-less formation shape control. 
The fundamental idea is to decompose a target formation into two parts: the first is the graphical formation shape and the second is its parameters including size, translation, and orientation in Euclidean space. 
The graphical formation shape is specified by the user and its parameters are generated automatically by our proposed algorithms. 

\subsection{Graphical Target Formation Shape}

Our approach requires a user-specified graphical shape as a starting point. 
In particular, the user can specify the target formation as a binary grid by the human-swarm interface we designed \cite{Sun2023HS}, where each cell is either black or white. 
The black cells correspond to the target shape. Robots are expected to move into the black cells such that the target formation could be achieved. 

Each cell is described by two parameters $\rho$ and $\xi_\rho$. 
Here, $\rho=(\rho_x,\rho_y)$ is the column and row indexes of a cell (see Fig.~\ref{Fig_overview}(b)). 
The scalar $\xi_\rho\in[0,1]$ is the color of the cell: $\xi_\rho=0$ if cell is black and $\xi_\rho=1$ if cell is white. 
Let $n_{\rm cell}$ be the number of black cells. 
Thus, the target shape can be described by the index set $\mathcal{F}=\{\rho: \xi_\rho=0\}$ and $|\mathcal{F}|=n_{\rm cell}$. 

Some important remarks about the graphical target formation are given below.
\begin{enumerate}[1)]
	\item Although the shape is discretized by a grid, robots still move continuously in space rather than jumping discretely from one cell to another. 
	The advantage of such an approach is that it greatly simplifies the shape design and meanwhile, respects the continuous dynamics of robots. 
	\item The graphical way provides a friendly and flexible interface to design the formation. 
	As a comparison, another possible way is to specify a formation shape by mathematical constraints. The disadvantage of this way is that the constraints would be extremely complex when the shape is complex.
	\item The graphical formation shape does not involve the parameters such as the size, translation, and rotation in the Euclidean space. As a result, users could easily specify the target shape in a graphical way without worrying about the physical parameters. As presented later, the parameters are generated automatically by our proposed algorithms. 
\end{enumerate}

Once the formation shape has been specified by the user, it is shared with all the robots. 
Each robot would have to further parameterize it so that it is implementable in the physical world. The parameterization process is given as bellow. 

\subsection{Size of the Target Formation}

The first parameter to be determined is the size of the formation shape. 
Since the target formation is represented by square cells, we only need to determine $l_{\rm cell}$, the size of cell. 

Let $r_{\rm avoid}$ be the collision-avoidance distance between two robots. It is expected the distance between each pair of robots in the target formation is equal to $r_{\rm avoid}$ (see Fig.~\ref{Fig_overview}(b)). 
As a result, the space occupied by each robot could be approximated by a regular hexagon with the center at the robot and the inscribed circle radius as $r_{\rm avoid}/2$. 
Thus, the area occupied by $n$ robots is $\frac{\sqrt{3}}{2}r_{\rm avoid}^2 n_{\rm robot}$. On the other hand, the area of the black cells is $l_{\rm cell}^2n_{\rm cell}$. The two areas are expected to be equal so that robots could fully occupy the formation $\frac{\sqrt{3}}{2} r_{\rm avoid}^2 n_{\rm robot}=l_{\rm cell}^2n_{\rm cell}$ from which $l_{\rm cell}$ can be solved as
\begin{align} \label{Equ_cellsizedef}
  l_{\rm cell} = \sqrt{\frac{\sqrt{3}}{2} \frac{ n_{\rm robot}}{n_{\rm cell}}}r_{\rm avoid} . 
\end{align}
Here, \eqref{Equ_cellsizedef} indicates that, when $n_{\rm robot}$ and $r_{\rm avoid}$ are given, $l_{\rm cell}$ is inversely proportional to $n_{\rm cell}$.

\begin{algorithm}[!t] \small \label{Alg_thread}
	\caption{Running Process.}
	\KwIn{$\mathcal{F}=\{\rho: \xi_\rho=0\}$, $n_{\rm robot}$}
	$l_{\rm cell}\leftarrow$ size parameterized by \eqref{Equ_cellsizedef} \\
	$\mathcal{F}_{\rm gray}\leftarrow$ gray conversion of $\mathcal{F}$ by \eqref{Equ_disttransform} \\
	\While{true} 
	{
		$msg\_{rec} \leftarrow$ all the messages received 
		\\
		$\{p_{\rm ref}, v_{\rm ref}, {\dot v}_{\rm ref}, \varphi_{\rm ref}, \omega_{\rm ref}, {\dot \omega}_{\rm ref}\} \leftarrow$ message in $msg\_{rec}$ 
		\\
		$\{p_{{\rm c},j}, v_{{\rm c},j}, {\dot v}_{{\rm c},j}, \varphi_{{\rm c},j}, \omega_{{\rm c},j}, {\dot \omega}_{{\rm c},j}\} \leftarrow$ message in $msg\_{rec}$ 
		\\  
		\If{stationary negotiation}
		{
			$p_{{\rm c},i}, \varphi_{{\rm c},i}\leftarrow$ negotiation update by \eqref{Equ_posesti_leaderless} and \eqref{Equ_headesti_leaderless} 
		}
		\Else
		{
			\If{$\{p_{\rm ref}, v_{\rm ref}, {\dot v}_{\rm ref}, \varphi_{\rm ref}, \omega_{\rm ref}, {\dot \omega}_{\rm ref}\} = \varnothing $}
			{
				$p_{{\rm c},i}, \varphi_{{\rm c},i}\leftarrow$ negotiation update by \eqref{Equ_posesti_leader_f} and \eqref{Equ_headesti_leader_f} 
			}
			\Else
			{
				$p_{{\rm c},i}, \varphi_{{\rm c},i}\leftarrow$ negotiation update by \eqref{Equ_posesti_leader_l} and \eqref{Equ_headesti_leader_l} 
			}
		}
		$\{p_j, v_j\} \leftarrow$ message in $msg\_{rec}$ \\
		$f_i^{\rm form} \leftarrow$ shape-forming force calculated by \eqref{Equ_forming} \\
		$f_i^{\rm stab} \leftarrow$ shape-stabilizing force calculated by \eqref{Equ_stabilizing} \\
		$f_i^{\rm avoid} \leftarrow$ collision-avoidance force calculated by \eqref{Equ_avoidance} \\
		$f_i^{\rm align} \leftarrow$ velocity-alignment force calculated by \eqref{Equ_alignment} \\
		$u_i \leftarrow f_i^{\rm form}+f_i^{\rm stab}+f_i^{\rm avoid}+f_i^{\rm align}$
		\\
		robot $i$ moves with command $u_i$ \\
		$msg\_{tra} \leftarrow\{p_{{\rm c},i}, v_{{\rm c},i}, {\dot v}_{{\rm c},i}, \varphi_{{\rm c},i}, \omega_{{\rm c},i}, {\dot \omega}_{{\rm c},i},p_i, v_i\}$
		\\
		transmit $msg\_{tra}$
	}
\end{algorithm}

\subsection{Position and Orientation of the Target Formation} 
\label{Subsec_posandhead}

Another two parameters of the target formation are the position and orientation. 
Our approach is to let the robots negotiate and reach a consensus in a distributed manner (see Fig.~\ref{Fig_overview}(c)). 
Specifically, each robot interprets the shape in terms of its local interest and initially conflicts with other robots. With the proposed protocols, each robot corrects the interpretation biases and reaches a consensus on the position and orientation of the final shape. 

Suppose $\rho_{\rm cen}$ is the cell that is located closest to the center of the shape. 
Let $p_{\rm cen}$ and $v_{\rm cen}$ be the position and velocity of the center point of cell $\rho_{\rm cen}$ in a global reference frame. 
Let $\varphi_{\rm cen}$ and $\omega_{\rm cen}$ denote the orientation angle and angular velocity of the target shape. 
Then, the position and orientation of the target shape could be represented by $p_{\rm cen}$ and $\varphi_{\rm cen}$, respectively. 
Here, we consider two scenarios depending on whether the target shape is required to track a time-varying reference (Algorithm~\ref{Alg_thread}, Lines 7-17). 

\paragraph{Stationary negotiation} 

In the stationary scenario, it is expected that the consensus value on $p_{\rm cen}$ and $\varphi_{\rm cen}$ is the mean of all the robots' initial interpretations, thus balancing the local interest of all the robots. 
Let $p_{{\rm c},i}$ and $v_{{\rm c},i}$ denote the interpretations of $p_{\rm cen}$ and $v_{\rm cen}$ by robot $i$. 
The interpretations of different robots can reach a consensus by 
\begin{align}
	\dot{v}_{{\rm c},i}=-\sum\limits_{\mathclap{j\in\mathcal{N}_i}}{a_{ij}(c_1 p_{{\rm c},ij}+c_2 v_{{\rm c},ij})}-v_{{\rm c},i}
	\label{Equ_posesti_leaderless}
\end{align}
where constant $c_1>0$, and variable $c_2=\mu_1+\frac{c_1}{\mu_1\gamma_i}$. 
Here, $\mu_1>0$ is a constant, and $\gamma_i$ is given by $\gamma_i=\sum\nolimits_{j\in\mathcal{N}_i}{a_{ij}}$. 
The relative bias of the interpretations of $p_{\rm cen}$ and $v_{\rm cen}$ are denoted as $p_{{\rm c},ij}=p_{{\rm c},i}-p_{{\rm c},j}$ and $v_{{\rm c},ij}=v_{{\rm c},i}-v_{{\rm c},j}$. 
Here, $a_{ij}\in \{0,1\}$ represents the adjacency coefficient. If $j$ is a neighbor of $i$, then $a_{ij}=a_{ji}=1$, and otherwise $a_{ij}=a_{ji}=0$. 
Initially, each robot interprets itself as the center of the target shape, i.e., $p_{{\rm c},i}(t_0)=p_i(t_0)$. It is locally optimal for the robots in the sense that they do not need to move since they are already within the target shape. 
There are two terms in \eqref{Equ_posesti_leaderless}. 
The first is the deficiency in the interpretations of $p_{\rm cen}$ and $v_{\rm cen}$ between robot $i$ and its neighbors, which aims to drive $p_{{\rm c},i}=p_{{\rm c},j}$ and $v_{{\rm c},i}=v_{{\rm c},j}$. 
The second term is the damping velocity, whose role is to drive $v_{{\rm c},i} \to 0$. 

Let $\varphi_{{\rm c},i}$ and $\omega_{{\rm c},i}$ be the interpretations of $\varphi_{\rm cen}$ and $\omega_{\rm cen}$ by robot $i$. 
The distributed negotiation on $\varphi_{\rm cen}$ is governed by 
\begin{align}
	\dot{\omega}_{{\rm c},i}=\sum\limits_{\mathclap{j\in\mathcal{N}_i}}{a_{ij}(c_3\varphi_{{\rm c},ij}+c_4\omega_{{\rm c},ij})}-\omega_{{\rm c},i}
	\label{Equ_headesti_leaderless}
\end{align}
where $c_3>0$ is a constant, and $c_4$ is defined as $c_4=\mu_2+\frac{c_3}{\mu_2\gamma_i}$. Here, $\mu_2>0$ is a constant. 
The relative bias $\varphi_{{\rm c},ij}$ and $\omega_{{\rm c},ij}$ are denoted as $\varphi_{{\rm c},ij}=\varphi_{{\rm c},i}-\varphi_{{\rm c},j}$ and $\omega_{{\rm c},ij}=\omega_{{\rm c},i}-\omega_{{\rm c},j}$. 
The initial value $\varphi_{{\rm c},i}(t_0)$ of the orientation interpretation could be randomly selected or follow the task-oriented requirements. 
As can be seen, the protocol in \eqref{Equ_headesti_leaderless} has the same structure as \eqref{Equ_posesti_leaderless} and can be analyzed analogously. 
In either $p_{\rm cen}$ or $\varphi_{\rm cen}$ negotiation, each robot only requires the neighbors' interpretations transmitted by wireless communication (Algorithm~\ref{Alg_thread}, Line 6), to reach a consensus on the final shape. 

Next, we show that the interpretations of $p_{\rm cen}$ and $\varphi_{\rm cen}$ converge to the means of all the robots' initial interpretations. 
In order to do that, we make the following assumption. 

\begin{assumption}
	The information graph $\mathcal{G}$ remains connected and time-invariant for all $t\geq 0$. 
	\label{Asm_connectedgraph}
\end{assumption}

Assumption~\ref{Asm_connectedgraph} is valid if the initial $\mathcal{G}$ is connected and the negotiation process converges much faster than the control one. 
As shown in Fig.~\ref{Fig_overview}(d), the proposed formation strategy consists of two processes: negotiation and control. Here, both of them are carried out simultaneously. 
In contrast to the control process, the negotiation is a virtual process and free of real robot's dynamic. As a result, the control process is much slower than the negotiation one. Simulation results support this deduction. In particular, when the negotiation converges, the swarm configuration remains unchanged. Thus, $\mathcal{G}$ is fixed and connected. 
In fact, a simple method that can strictly ensure Assumption~\ref{Asm_connectedgraph} would be to run an initialization process during which we first wait for the negotiation process to converge. 

Then, we first discuss the nonsingularity of $\gamma_i$ by Lemma~\ref{Thm_nonsigularity} and then show the convergence of \eqref{Equ_posesti_leaderless} and \eqref{Equ_headesti_leaderless} by Theorem~\ref{Thm_estimate_leaderless}.

\begin{lemma}
	Under Assumption~\ref{Asm_connectedgraph}, $\gamma_i>0$ for all $i \in \mathcal{V}$. 
	\label{Thm_nonsigularity}
\end{lemma}
\begin{proof}
	See the Appendix. 
\end{proof}

\begin{theorem}
	Suppose $v_{{\rm c},i}(t_0)=0$, $\omega_{{\rm c},i}(t_0)=0$. 
	If $\mathcal{G}$ is fixed and connected, under protocols \eqref{Equ_posesti_leaderless} and \eqref{Equ_headesti_leaderless}, the interpretations of $p_{\rm cen}$ and $\varphi_{\rm cen}$ converge to the means of all the robots' initial interpretations globally and asymptotically. 
	\label{Thm_estimate_leaderless}
\end{theorem}
\begin{proof}
	See the Appendix. 
\end{proof}

\paragraph{Time-varying negotiation} 

To handle maneuvering formations, we introduce a small number of informed robots who know the global reference (Algorithm~\ref{Alg_thread}, Line 5). 
The rest uninformed robots negotiate the shape towards the informed ones based on the local interpretations of neighboring robots (Algorithm~\ref{Alg_thread}, Line 6). 

In particular, uninformed robots apply a variant of \eqref{Equ_posesti_leaderless} to update their position interpretations as follows: 
\begin{subequations} 
	\begin{align}
		\dot{v}_{{\rm c},i}=-\sum\limits_{j\in\mathcal{N}_i}{a_{ij}(c_1 p_{{\rm c},ij}+c_2 v_{{\rm c},ij})}+\dfrac{1}{\gamma_i}\sum\limits_{j\in\mathcal{N}_i}{a_{ij}\dot{v}_{{\rm c},j}}
		\label{Equ_posesti_leader_f}
	\end{align}
	\label{Equ_posesti_leader}
\end{subequations}
and informed robots execute the following protocol: 
\begin{align}
	\begin{split}
		\dot{v}_{{\rm c},i}
		=&-\sum\limits_{j\in\mathcal{N}_i}{a_{ij}(c_1 p_{{\rm c},ij}+c_2 v_{{\rm c},ij})}-[c_1(p_{{\rm c},i}-p_{\rm ref})\\
		&+c_2(v_{{\rm c},i}-v_{\rm ref})-\dot{v}_{\rm ref}]
	\end{split}
	\tag{\ref{Equ_posesti_leader}b}
	\label{Equ_posesti_leader_l}
\end{align}
where $p_{\rm ref}, v_{\rm ref} \in \mathbb{R}^d$ are the position and velocity of the time-varying reference. 
Here, the reference signal is specified by the user, such as through trajectory planning (Algorithm~\ref{Alg_thread}, Line 5). 
The role of the first term in \eqref{Equ_posesti_leader} is the same as \eqref{Equ_posesti_leaderless}. 
The second term in \eqref{Equ_posesti_leader_f} is the average of the velocities of the neighbors. 
Its role is to drive $v_{{\rm c},i}\to v_{{\rm c},j}$. 
The second term in \eqref{Equ_posesti_leader_l} aims to track the reference, that is, $p_{{\rm c},i}\to p_{\rm ref}$ and $v_{{\rm c},i}\to v_{\rm ref}$. 

Regarding the orientation negotiation, uninformed robots execute the following protocol:
\begin{subequations} 
	\begin{align}
		\dot{\omega}_{{\rm c},i}=-\sum\limits_{j\in\mathcal{N}_i}{a_{ij}(c_3\varphi_{{\rm c},ij}+c_4\omega_{{\rm c},ij})}+\dfrac{1}{\gamma_i}\sum\limits_{j\in\mathcal{N}_i}{a_{ij}\dot{\omega}_{{\rm c},j}}
		\label{Equ_headesti_leader_f}
	\end{align}
	\label{Equ_headesti_leader}
\end{subequations}
and informed robots use the following protocol: 
\begin{align}
	\begin{split}
		\dot{\omega}_{{\rm c},i}
		=&-\sum\limits_{j\in\mathcal{N}_i}{a_{ij}(c_3\varphi_{{\rm c},ij}+c_4\omega_{{\rm c},ij})}-[c_3(\varphi_{{\rm c},i}-\varphi_{\rm ref})\\
		&+c_4( \omega_{{\rm c},i}-\omega_{\rm ref})-\dot{\omega}_{\rm ref}]
	\end{split}
	\tag{\ref{Equ_headesti_leader}b}
	\label{Equ_headesti_leader_l}
\end{align}
where $\varphi_{\rm ref}, \omega_{\rm ref} \in \mathbb{R}^d$ are the orientation and angular velocity of the time-varying reference. 
The protocol in \eqref{Equ_headesti_leader} has the same structure as \eqref{Equ_posesti_leader} and can be analyzed approximately. 

Next, we show the convergence of the proposed negotiation protocols \eqref{Equ_posesti_leader} and \eqref{Equ_headesti_leader} by Theorem~\ref{Thm_estimate_leader}. 

\begin{theorem}
	If $\mathcal{G}$ is fixed and connected, under protocols \eqref{Equ_posesti_leader} and \eqref{Equ_headesti_leader}, the interpretations of $p_{\rm cen}$ and $\varphi_{\rm cen}$ converge to the time-varying reference globally and asymptotically. 
	\label{Thm_estimate_leader}
\end{theorem}
\begin{proof}
	See the Appendix. 
\end{proof}

\section{Distributed Controller Design}
\label{Sec_controller}

Previously, we have defined the target formation, and then we present a distributed control law (Algorithm~\ref{Alg_thread}, Lines 18-28). 
The implementation structure is illustrated in Fig.~\ref{Fig_overview}(d). 
First, the graphical shape is sent to all the robots. Then, each robot converts a graphical shape into a gray grid. Next, robots negotiate $p_{\rm cen}$ and $\varphi_{\rm cen}$ of the target formation through the consensus protocols. 
To implement identity-less shape formations, we still need to solve the problem of how to design interaction behaviors using only locally sensed information. 

\subsection{Gray conversion of the target formation}

Given a target formation specified by a binary grid, the first step to achieve such a formation is to convert the binary grid to a gray one (Algorithm~\ref{Alg_thread}, Line 2). 
By doing so, we could expand the influence of the target formation so that each robot can move into the formation smoothly along the grayscale. 

\begin{figure}[!t]
	\centering
	\includegraphics[width=8.5cm]{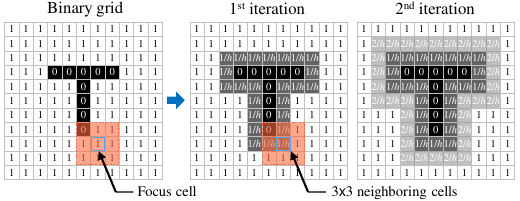}
	\caption{An example to illustrate the gray-conversion process of \eqref{Equ_disttransform}.}
	\label{Fig_conversion}
\end{figure}

We implement the gray conversion of the binary grid by introducing the idea of \emph{distance transformation}. Specifically, it takes the cell as the operating unit, and expands the binary grid out by~$h$ cells based on the black cells to form an~$h$-level gray grid. 
To obtain the gray value~$\xi_{\rho}$ of cell $\rho$, we use a local parallel method described in~{\cite{Borgefors1986Gray}} as follows: 
\begin{equation}
  \xi_{\rho}^k = \min \limits_{\rho' \in \mathcal{M}_{\rho}} {\left({\xi_{\rho'}^{k - 1} + \dfrac{1}{h}}\right)}
  \label{Equ_disttransform}
\end{equation}
where the superscript $k$ represents the $k$-th iteration, and $\mathcal{M}_{\rho}$ is composed of~$3 \times 3$ cells including~$\rho$ and its surrounding cells (see~Fig.~\ref{Fig_conversion}). 
With iteration \eqref{Equ_disttransform}, the gray formation shape is $\mathcal{F}_{\rm gray}=\{\rho: \xi_{\rho} \in [0,1)\}$. 
It is obvious that $\mathcal{F} \subseteq \mathcal{F}_{\rm gray}$. 

Then we use an example to explain the iteration process in \eqref{Equ_disttransform}, as shown in Fig.~\ref{Fig_conversion}. 
An original binary grid, to which the distance transformation is to be applied, consists of black cells with the initial value of zero, and white cells with the initial value of one. See the left diagram in Fig.~\ref{Fig_conversion}. 
At each iteration, each cell only considers the neighboring cells within its local~$\mathcal{M}_{\rho}$, and selects the smallest distance value as the new gray color of the cell. 
The other two diagrams in Fig.~\ref{Fig_conversion} describe the distance changes of the grid image at the first and second iterations, respectively. The process is repeated until no cell value changes, that is, the number of iterations is proportional to the gray scale $h$. 

\subsection{Distributed Formation Shape Control Law}
\label{Subsec_formcontrol}

To implement dentity-less shape formations, we design a distributed control law
\begin{align*} 
  u_i=f_i^{\rm form}+f_i^{\rm stab}+f_i^{\rm avoid}+f_i^{\rm align}
\end{align*}
where $f_i^{\rm form}$, $f_i^{\rm stab}$, $f_i^{\rm avoid}$, and $f_i^{\rm align}$ represent the shape-forming, shape-stabilizing, collision-avoidance, and velocity-alignment forces, respectively. 
Note that the calculation of these forces only requires neighbors' positions and velocities which can be obtained by wireless communication (Algorithm~\ref{Alg_thread}, Line 18). 
In particular, robot's own position and velocity require obtaining by an external positioning system. 

\paragraph{Shape-forming force} 

The purpose of shape-forming force~$f_{i}^{\rm form}$ is to drive robots into the target formation along the gray gradient (Fig.~\ref{Fig_overview}(e)). The force~$f_{i}^{\rm form}$ is defined as
\begin{align} 
	f_i^{\rm form}=\kappa_1(1-\cos{\pi\xi_{\rho,i}})g_{ik}
	\label{Equ_forming}
\end{align}
where $\kappa_1$ is a positive constant, and $g_{ik}=\frac{p_{k,i}-p_i}{\|p_{k,i}-p_i\|}$ is a unit vector pointing from $p_i$ to $p_{k,i}$. 
Here, $g_{ik}$ points to the descending direction of the gray scale~$\xi_{\rho,i}$, and $p_{k,i}$ is the center position of the darkest cell $\rho_k$ around robot $i$. 
The value $\xi_{\rho,i}$ indicates the gray color of the corresponding cell $\rho_i$ of robot $i$ in the gray shape. 

The states $\xi_{\rho,i}$ and $p_{k,i}$ are critical for $f_{i}^{\rm form}$ to realize identity-less formation. 
Before calculating these two states, the robot first needs to locate itself in the shape grid by $\rho_i=\left\lceil{R_{\varphi_{{\rm c},i}}^{-1} (p_i - p_{{\rm c},i})}/{l_{\rm cell}}\right\rfloor+\rho_{\rm cen}$, where $\left\lceil\cdot\right\rfloor$ represents rounding to the nearest integer, and $R_{\varphi_{{\rm c},i}}$ is defined as $R_{\varphi_{{\rm c},i}}=\left[\cos\varphi_{{\rm c},i}, -\sin\varphi_{{\rm c},i}; \sin\varphi_{{\rm c},i}, \cos\varphi_{{\rm c},i} \right]$. 
Then, $\xi_{\rho,i}$ is determined by the color of $\rho_i$ in the gray shape $\mathcal{F}_{\rm gray}$. If $\rho_i$ is outside $\mathcal{F}_{\rm gray}$, then $\xi_{\rho,i}=1$. 
Next, the robot can calculate $p_{k,i}$ by $p_{k,i}=R_{\varphi_{{\rm c},i}} (\rho_k - \rho_{\rm cen}) l_{\rm cell}+p_{{\rm c},i}$, where $\rho_k$ is denoted as the local target cell, i.e., the darkest cell around robot $i$. 
Here, we need to consider two cases. 
The first is $\rho_i\in\mathcal{F}_{\rm gray}$, that is $\xi_{\rho}<1$. 
In this case, robot $i$ can select a cell with the smallest gray color around its location $\rho_i$ as $\rho_k$, that is, $\rho_k=\mathop{\arg\min}\nolimits_{\rho \in \mathcal{C}_{\rho,i}} {\xi_{\rho}}, \xi_{\rho} < \xi_{\rho_i}$, where $\mathcal{C}_{\rho,i}$ is the set of neighboring cells within $r_{\rm sense}$ around $\rho_i$. 
For the second case where $\rho_i\notin\mathcal{F}_{\rm gray}$, namely $\xi_{\rho}=1$, robot $i$ can choose the nearest gray cell as $\rho_k$, that is, $\rho_k=\mathop {\arg \min }\nolimits_{\rho \in \mathcal{F}_{\rm gray}} \| p_i - p_{\rho} \|, \xi_{\rho} < 1$, 
where $p_{\rho}$ is the center position of cell $\rho$ and can be obtained by a calculation similar to $p_{k,i}$. 

As can be seen, if robot $i$ is outside the gray formation ($\xi_{\rho,i}=1$), it will enter the gray formation as soon as possible. If robot $i$ is in the gray formation ($\xi_{\rho,i}\in (0,1)$), it will fall into the target formation along the descending direction of $\xi_{\rho,i}$. If robot $i$ enters the target formation ($\xi_{\rho,i}=0$), then it is no longer affected by $f_{i}^{\rm form}$. 

\paragraph{Shape-stabilizing force} 

As mentioned before, $f_{i}^{\rm form}$ mainly works when robot $i$ is outside the target formation. For this reason, we design the shape-stabilizing velocity $f_{i}^{\rm stab}$ acting within the formation shape, to stabilize robots in the target formation (Fig.~\ref{Fig_overview}(e)). 
The force $f_{i}^{\rm stab}$ can be defined as
\begin{align} 
	f_i^{\rm stab}=\kappa_2\left(\dfrac{1}{m_i} {\sum\limits_{\rho \in \mathcal{B}_{\rho,i}} p_{\rho}} - p_{i}\right)
	\label{Equ_stabilizing}
\end{align}
where constant $\kappa_2>0$, and $m_i=\left|{\mathcal{B}_{\rho,i}}\right|$. 
The set $\mathcal{B}_{\rho,i}$ is composed of all the black cells within $r_{\rm avoid}$ around robot $i$. 
As seen from \eqref{Equ_stabilizing}, the robot selects the geometric center of all the black cells in its surrounding area as the local target (see Fig.~\ref{Fig_overview}(e)). 
Then, the robot controls the position deviation from the local target to stabilize itself in the target formation. 

As mentioned before, $f_{i}^{\rm form}$ mainly works when robot $i$ is outside the target formation. 
However, due to the lack of control within the formation shape, robots would stop moving near the shape boundary, and even robots that have entered the formation may leave again. 
Fortunately, $f_i^{\rm stab}$ solves the above problems and adds the control influence while the robot entering the target shape. 
Moreover, it can ensure that all the robots entering the target formation match with a certain cell in the formation shape. 
If there are both black and white cells around robot $i$, $f_i^{\rm stab}$ drives it to a black cell far away from the shape boundary to avoid leaving the target shape. 
If all the cells around robot $i$ are black, then $f_i^{\rm stab}$ controls it to slide inside the formation according to the actions of its neighbors. 
In a word, $f_i^{\rm stab}$ can prevent a robot from leaving the target formation, but it does not affect the movement of the robot in the formation. 

\paragraph{Collision-avoidance force} 

Whether inside or outside the formation, robots may collide with their neighbors. For this reason, we define the collision-avoidance force $f_i^{\rm avoid}$ to drive robots to avoid collisions with their neighbors (Fig.~\ref{Fig_overview}(e)). 
In addition, $f_{i}^{\rm avoid}$ also affects the distribution density of robots in the target formation, and it can work with~$f_{i}^{\rm stab}$ to make all the robots in the swarm evenly distributed in the target formation. 
The control force~$f_{i}^{\rm avoid}$ is given by
\begin{align} 
	f_i^{\rm avoid}=-\kappa_3\sum\limits_{j\in\mathcal{N}_i}\psi\left(\dfrac{\|p_j-p_i\|}{r_{\rm avoid}}\right)g_{ij}
	\label{Equ_avoidance}
\end{align}
where $\kappa_3 >0$ is a constant, and $g_{ij}=\frac{p_j-p_i}{\|p_j-p_i\|}$ is a unit vector. 
Recall that $r_{\rm avoid}$ is the collision-avoidance scope of robots. 
The function $\psi(z)$ in \eqref{Equ_avoidance} is defined as (6) in \cite{Li2024RAL}. 
This function is monotonically decreasing from 1 to 0 as $z$ increases. 
As a result, the collision-avoidance effect between robots $i$ and $j$ decrease as $\|p_j-p_i\|$ increase. 
If $\|p_j-p_i\|$ is close to $r_{\rm avoid}$, the collision-avoidance effect monotonically decreases to zero. 

The control forces $f_{i}^{\rm stab}$ and $f_{i}^{\rm avoid}$ jointly affect the uniform distribution of robots in the target formation. 
The role of $f_{i}^{\rm stab}$ is to stabilize robots in the formation shape and prevent them from leaving the formation again. 
The objective of $f_{i}^{\rm avoid}$ is to make all the robots within the formation evenly distributed in the formation shape through repulsion. 
It should be pointed out that the space size of the target formation is compatible with the group scale as discussed in~\eqref{Equ_cellsizedef}. 
Therefore, by the coordination of~$f_{i}^{\rm stab}$ and~$f_{i}^{\rm avoid}$, robots can be evenly distributed in the target formation.

\paragraph{Velocity-alignment force} 

Finally, in order to realize the coordination movement among robots, we introduce the control component~$f_{i}^{\rm align}$, which is defined as
\begin{align} 
	f_i^{\rm align}=-\sum\limits_{j\in\mathcal{N}_i}{a_{ij}\left(v_i-v_j\right)}+\left[\kappa_4\left(\hat v_{{\rm c},i}-v_i\right)+\dot{\hat v}_{{\rm c},i}\right]
	\label{Equ_alignment}
\end{align}
where $\kappa_4$ is a positive constant. Recall that $a_{ij}\in\{0,1\}$ is the adjacency coefficient. 

The first term in \eqref{Equ_alignment} aims to align robots' velocities with their neighbors. 
Since the evolution direction of this term is opposite to the direction of $v_i-v_j$, it will reduce the difference of $v_i-v_j$ until $v_i$ and $v_j$ reach a consensus. 
Thus, the first term also acts as a friction force. 
The second term in \eqref{Equ_alignment} aims to align robot $i$'s velocity with its local interpretation on the moving velocity of the target formation. 
This term is necessary when the target formation is maneuvering. 

Up to now, we have solved the problem raised at the beginning of this section. 
As seen from \eqref{Equ_forming}-\eqref{Equ_alignment}, no label is introduced to identify and match between robots and the target formation. 
In contrast, robots achieve the target formation by covering the user-specified shape area, rather than moving to the uniquely assigned goal locations in the shape. 
Here, each robot only needs to determine the local cells and corresponding gray colors, so as to form the shape. 

Some meaningful discussions about the proposed control strategy are given as below.
\begin{enumerate}[1)]
	\item The target formation is required to be a single connected shape. This is because the designed behaviors are based on local information. If two disconnected shapes are far away, robots lack the ability to sense the cells across different shapes. This problem may be addressed by preassigning robots to different shapes initially. 
	\item The global position of each robot is required in the process of formation control. With this global information, robot can evaluate its location relative to the target formation. In practice, robot can use GPS to obtain its global positions outdoors or a motion capture system to obtain it indoors. 
	\item The negotiation process does not rely on external positioning systems and is thus not affected by positioning noise. The reason is that negotiation is a purely numerical process. For the control process, the final shape would be deformed in the presence of positioning noise, which is a common phenomenon for cooperative control tasks \cite{Sun2023Aseembly}. 
\end{enumerate}

\section{Simulation and Experiment Results}
\label{Sec_results} 

To evaluate the performance of our proposed strategy, we implement and test our proposed strategy in both simulation and experiment. 
Note that the metric definitions of entering rate and uniformity refer to \cite{Sun2023Aseembly}. The convergence time is defined as the time when the entering rate is equal to 100\%. 

\subsection{Simulation Results}
\label{Subsec_simulation} 

\begin{figure*}[!t]
	\centering
	\includegraphics[width=17.9cm]{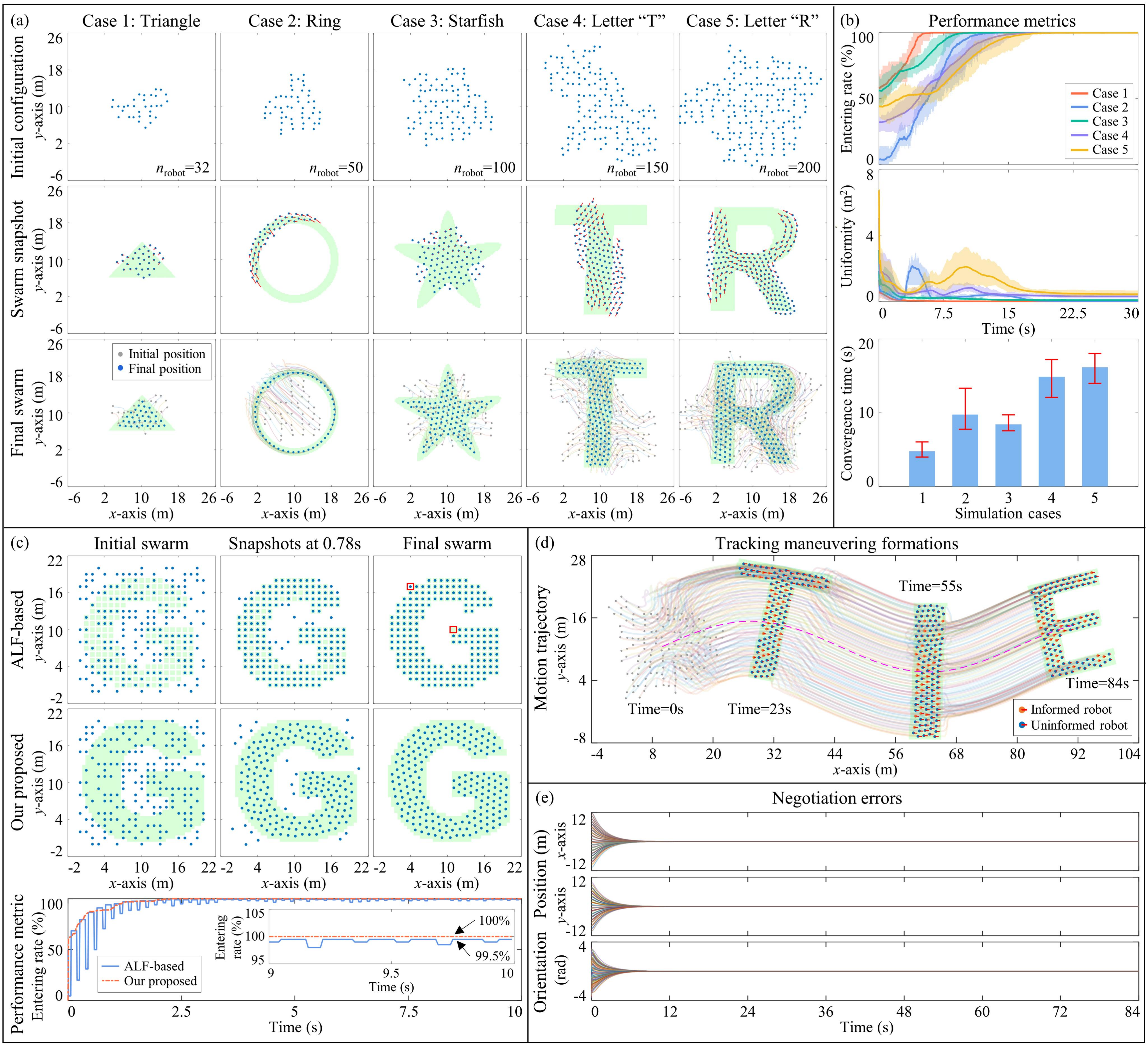}
	\caption{Simulation results. 
	(a)~Snapshots and trajectories of shape formation processes under different shapes. 
	(b)~Statistic results of 100 trials in (a). 
	(c)~Comparison results between our proposed strategy and the artificial-light-field (ALF) based method in \cite{Chu2023TASE}. 
	(d)~Swarm trajectories of the maneuvering formation process. 
	(e)~Negotiation errors of the formation position and orientation with respect to the time-varying reference in (d).}
	\label{Fig_simulation}
\end{figure*}

We present three simulation examples to test and verify the proposed strategy, as illustrated in Fig.~\ref{Fig_simulation}. 
Here, robot is modeled as a circular robot with a body size of $r_{\rm body}=0.2{\rm m}$. 
Each robot is able to sense any other robot who lies within its sensing range $r_{\rm sense}=2.5{\rm m}$. 
Furthermore, collision avoidance is triggered if the inter-robot distance is less than $r_{\rm avoid} = 1.5{\rm m}$. 
Other parameters are listed as: $c_1=1.5, \mu_1=3.0, c_3=0.8, \mu_2=1.5, \kappa_1=30.0, \kappa_2=100.0, \kappa_3=80.0, \kappa_4=1.0$. 

\paragraph{Effectiveness for complex shape} 

The first example is used to investigate the effectiveness of our proposed strategy. 
In this example, we present five cases as shown in Fig.~\ref{Fig_simulation}(a), which correspond to five formations with different patterns and $n_{\rm robot}$. 
In each case, we all present $20$ trails with a random initialization, and for a total of $100$ trails. 
Fig.~\ref{Fig_simulation}(b) shows the statistical metrics of all $100$ trails in Fig.~\ref{Fig_simulation}(a). 
We first present a simple case where the target shape is a convex triangle with $n_{\rm robot}=32$. 
As shown in Fig.~\ref{Fig_simulation}(a) and (b), all the robots smoothly form the triangle formation, and fully enter the target formation by $3.99{\rm s}$ to $6.05{\rm s}$ in all $20$ trails. 
Moreover, robots are almost evenly distributed in the triangle formation because the uniformity is stable at $0.05$. 
Then, we present the second case where a swarm of $50$ robots form a concave ring shape. 
As shown in Fig.~\ref{Fig_simulation}(a), our proposed strategy can also assemble the concave ring shape and realize uniform distribution verified by the fact that the entering rate converges to $100\%$ and the uniformity is stable at $0.04$. 
In addition to the above two cases, our strategy can form more complex shapes such as starfish and letter shapes, and adapt to large-scale robot swarms, as demonstrated in Fig.~\ref{Fig_simulation}(a) and (b). 
It is shown by the statistical results in Fig.~\ref{Fig_simulation}(b) that our strategy exhibits stable performance in different shapes and swarm sizes. 
Specifically, the entering rate remains $100\%$ and the uniformity converges to a smaller value. 
Moreover, as the group scale increases from $32$ to $200$, the convergence time increases mildly, which is verified by Fig.~\ref{Fig_simulation}(b).  

In the second example, the effectiveness of our proposed strategy is compared with the state-of-the-art method in \cite{Chu2023TASE}, called artificial-light-field (ALF) based method, as shown in Fig.~\ref{Fig_simulation}(c). 
The ALF is also an assignment-based method. 
In the comparison, a group of $200$ robots forms a letter shape ``G''. 
To ensure a fair comparison, we set that all the robots know the location of the target shape and the movement speed of each robot is limited to $3m/s$. 
It is worth emphasizing that the inter-robot collision avoidance is not considered in ALF-based method, which is consistent with \cite{Chu2023TASE}. 
The comparison result is shown in Fig.~\ref{Fig_simulation}(c).  
It is shown that, given the same initial configurations, the two methods converge towards the target formation at approximately convergence speed. 
In contrast to our method, the entering rate of ALF-based method oscillates repeatedly, since the robot needs to move between the discretized points when reassigning the goal locations, as shown in Fig.~\ref{Fig_simulation}(c). 
Above all, ALF-based method cannot form the target shape due to the fact that goal locations are not fully occupied by the robots, as marked by the red box in Fig.~\ref{Fig_simulation}(c). 
This is also reflected in the entering rate where ALF-based method only reaches $99.5\%$ in the end while our method is $100\%$. 
The reason is that robots easily fall into local minima, also known as \emph{deadlock}, based only on local measurements \cite{Sun2023Aseembly}. 
In contrast, our method can successfully form the target shape without getting deadlock anywhere. 

\paragraph{Tracking maneuvering formation} 

Next, we present the third example to test the maneuverability of our proposed control strategy in terms of translation, rotation, and even shape deformation. 
See Fig.~\ref{Fig_simulation}(d) for an illustration. 
In this task, a swarm of $128$ robots sequentially forms three letter shapes, namely, ``T", ``I", and ``E", from random initialization. 
In this example, only one in five robots in the swarm knows the information of the reference who moves in a sinusoidal motion. 
Note that the reference signal is specified by the user in real time through wireless communication. 
The reference trajectory is planned in advance. 
As shown in Fig.~\ref{Fig_simulation}(d), our proposed strategy has enough ability to drive a robot swarm to implement both maneuvering formation and shape deformation. 
All robots' interpretations of the formation position and orientation converge to the time-varying reference in about $10{\rm s}$, as demonstrated in Fig.~\ref{Fig_simulation}(e). 
Moreover, once a consensus is reached, it remains constant regardless of the robots' movements and even shape deformation. 
This is mainly due to the fact that the negotiation loop is decoupled from the control loop as depicted in Fig.~\ref{Fig_overview}(e). 

\paragraph{Parameter settings} 

There is a total of 11 parameters in our proposed strategy, that is, $r_{\rm body}$, $r_{\rm sense}$, $r_{\rm avoid}$, $c_1$, $c_3$, $\mu_1$, $\mu_2$, $\kappa_1$, $\kappa_2$, $\kappa_3$, and $\kappa_4$. 
The first three parameters are robot-fixed because they are constrained by the robot's own capabilities. 
The middle four parameters determine the negotiation process. 
They are insensitive in both simulation and experiment due to the fact that negotiation is a virtual process and free of real robot's dynamics. 
The last four parameters determine the shape control process and their settings are crucial to the performance as mentioned in \cite{Jin2022IFAC}. 
These parameters are manually adjusted in the order of $\kappa_1$, $\kappa_3$, $\kappa_2$, and $\kappa_4$. 
More discussion is given in Appendix. 
The parameter settings of all three simulation examples are consistent. This implies that the parameters are insensitive to shape changes and motion. 
If the robot's dynamics and parameters ($r_{\rm avoid}$ and $r_{\rm sense}$) change, the control parameters ($\kappa_1$, $\kappa_2$, $\kappa_3$, and $\kappa_4$) need to be readjusted, verified by the following experiments. 

\begin{figure*}[!t]
	\centering
	\includegraphics[width=18.1cm]{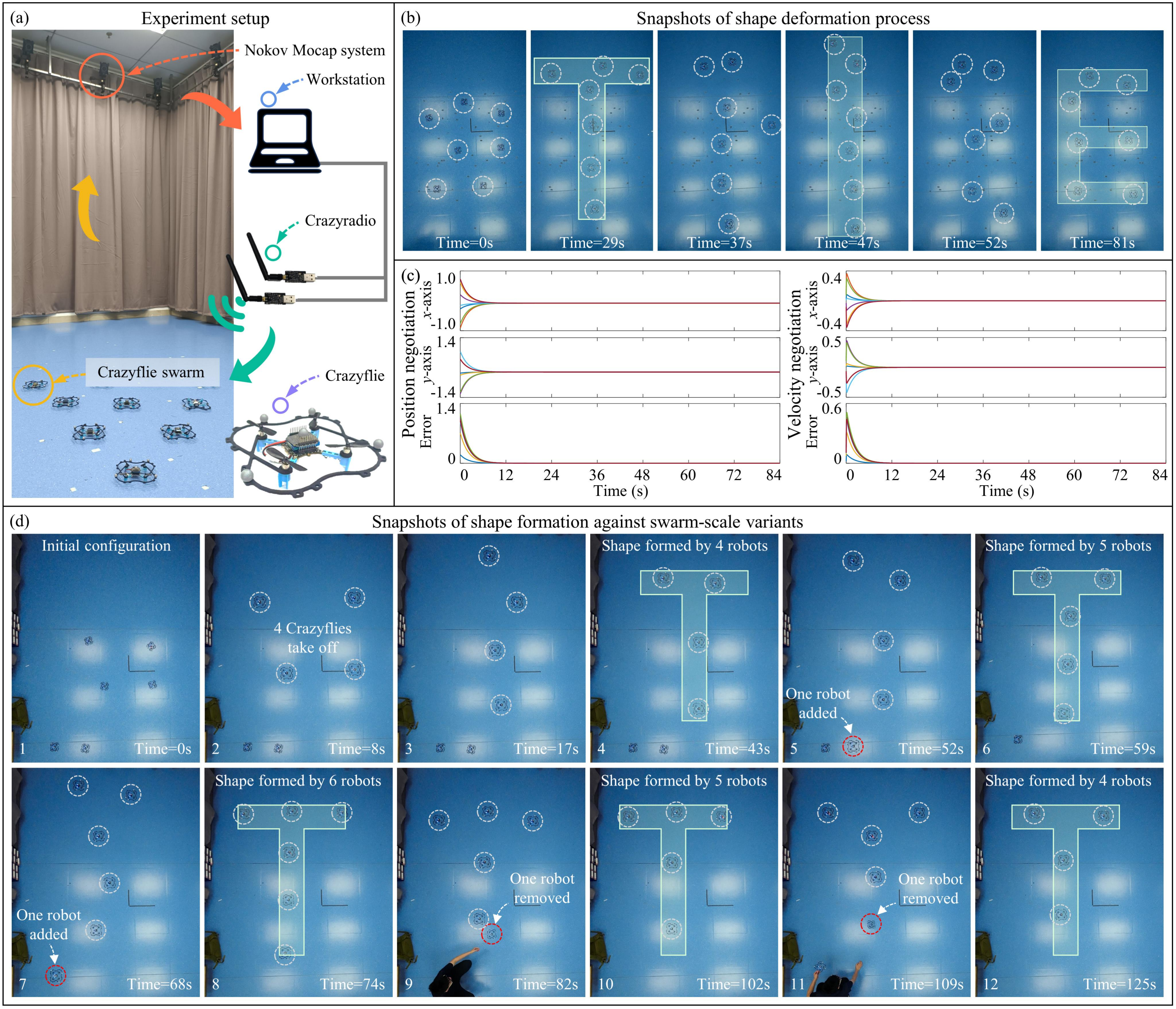}
	\caption{Experiment results. 
	(a)~Experiment and implementation setup. 
	(b)~Snapshots of $7$ Crazyflies forming three different shapes in a sequence. 
	(c)~Position and velocity negotiation in (b). 
	(d)~Snapshots of shape formation against group-scale changes. 
	Initially, 4 Crazyflies form the target shape. Then, as 2 Crazyflies are added to and removed from the swarm one by one, the target shape is still formed. }
	\label{Fig_experiment}
\end{figure*}

\subsection{Experiment Results}
\label{Subsec_experiment} 

To validate the correctness of the proposed control strategy beyond simulation, we implement experiment demonstration (See Fig.~\ref{Fig_experiment}) using the Crazychoir swarm system \cite{Pichierri2023Crazychoir}, an open-source flying platform that allows running experiments on a swarm of Crazyflie nano-robots. 

\paragraph{Experimental setup} 

There are four key components in the experiment, as shown in Fig.~\ref{Fig_experiment}(a). 
The first is a group of $7$ Crazyflies used to implement formation shape control. Each robot can receive messages through a $2.4{\rm GHz}$ Crazyradio. All the robots in the swarm are homogeneous and have no identities. 
The second is a workstation equipped with the Crazychoir for controlling all the robots. 
In Crazychoir, each robot is handled by a standalone process (also called ROS~2 \emph{node}), as shown in Fig.~\ref{Fig_overview}(d). 
The third is the Crazyradio dongle used to establish radio communication between robots and their nodes in the Crazychoir, as shown in Fig.~\ref{Fig_experiment}(a). 
Here, we use two Crazyradios to avoid channel congestion and reduce transmission latency. 
The fourth is a Nokov Mocap system used to acquire the position of each robot at a rate of $200{\rm Hz}$. The linear velocity of robot is obtained by taking the derivative and low-pass filtering of its position. 
Crazychoir provides an interface for converting acceleration commands $u_i$ into attitude signals that Crazyflie can receive. The detailed conversion process is shown in the Appendix. 

\paragraph{Implementation Using Crazyflies} 

We demonstrate the proposed strategy by two real experiments. 
The first is a group of $7$ robots forming three different letter shapes of ``T'', ``I'', and ``E'', in a sequence, as shown in Fig.~\ref{Fig_experiment}(b). 
The parameters are chosen as: $r_{\rm sense}=2.5{\rm m}, r_{\rm avoid}=0.9{\rm m}, \kappa_1=1.0, \kappa_2=8.0, \kappa_3=6.0$. Others are the same as in simulation. 
In the experiment, the proposed strategy exhibits smooth swarming motion in shape forming and deforming tasks. 
As shown in the experiment result in Fig.~\ref{Fig_experiment}(b), the robot swarm can form three letter patterns and switch from one to another smoothly, which benefits from no requirement of unique identities and independence of assignment of our proposed strategy. 
Moreover, the distributed negotiation process among all the robots converges in about 10${\rm s}$, as demonstrated in Fig.~\ref{Fig_experiment}(c). 
The experiment video can be found in the supplementary file. 

\paragraph{Robustness to swarm variants} 

The second experiment is used to verify the robustness to the group-scale variants. 
Fig.~\ref{Fig_experiment}(d) shows the experiment result where the target pattern is a letter shape ``T''. 
The parameters applied are the same as the first experiment. 
Initially, a group of $4$ Crazyflies takes off and successfully forms the target shape (the first four snapshots in Fig.~\ref{Fig_experiment}(d)). 
Then, we add one robot to the swarm. 
The new swarm can still form the target shape, as shown in snapshots 5 and 6 in Fig.~\ref{Fig_experiment}(d). 
Similarly, we add one more robot to the swarm after the robots achieves the formation. 
It is shown from snapshots 7 and 8 in Fig.~\ref{Fig_experiment}(d) that a new swarm of 6 robots can also successfully form the target shape. 
The newly formed letter ``T'' has more number of robots than the original one, which demonstrates the robustness of our proposed strategy against newly added robots. 
The robustness is due to the identify-less control strategy, by which robots can replace the roles with each other. 
Next, we remove the two robots one by one, the rest of the swarm spontaneously form the target shape, as shown in the last four snapshots in Fig.~\ref{Fig_experiment}(d). 
It is also shown that our proposed strategy has robustness against robot failures. 
Specifically, the robots can replace the roles of the removed robots and successfully reform the shape. 

Notably, each new robot joining will affect the robots' interpretations of the final shape since those interpretations of the newly added robot deviate from that of the original robots. 
This is the reason why the shape position moves towards the new robot each time a new robot is added (snapshots 4, 6, and 8 in Fig.~\ref{Fig_experiment}(d)). 
The shape orientation never changes since we set each robot's initial interpretation on shape orientation to $0$ in the experiment. 
Different from adding robots, robots' interpretations of the final shape are free of removing robots (snapshots 8, 10, and 12 in Fig.~\ref{Fig_experiment}(d)) since they are not governed by robots' movements. 
 
\section{Conclusion}
\label{Sec_conclusion}

In this paper, we proposed an identity-less distributed shape formation strategy that drove a swarm of homogeneous robots starting from any given initial configuration to move and form a user-specified shape smoothly. 
With the proposed strategy, robots took a user-specified graphical shape as a starting point, then used local information to negotiate the Euclidean parameters of the target formation and assemble the target formation shape concurrently. 
Moreover, our proposed strategy can also achieve formation maneuver control such as translation, rotation, or even shape deformation of the target formation. 
To evaluate the correctness and performance of our proposed strategy, we executed our strategy on a swarm of up to $200$ simulated robots and $7$ aerial robots. 
The results showed that our strategy exhibits high efficiency and stable performance in different shapes and group sizes, and the convergence time increases mildly as the swarm size increases. 

The shape formation considered in this paper can be viewed as a full cover of the target grid by robots, due to the grid representation of the formation shape, in the sense that it is similar to the coverage control problem. 
In comparison, the difference between the two is also obvious. 
First, coverage control focuses on the cover of hot areas usually represented by Gaussian fields, rather than forming complex shapes. 
Second, the coverage problem needs to construct a coverage-rate function and find its gradient to yield the control law. 
This is essentially similar to the traditional artificial-potential-field method, which may easily get trapped in local minima when forming complex shapes. It would be meaningful to illustrate the substantial differences between the two by using quantitative experiments in the future. 

There are several interesting avenues for future work. For example, the formation strategy proposed in this paper can be generalized by considering measurement noise, obstacle avoidance, communication delay, sensing-range variation, nonholonomic dynamics, and connectivity preservation, which are currently impossible. 
In particular, it is also necessary to discuss the proposed strategy's performance variations under the above non-ideal conditions. 
Secondly, it is meaningful to present the theoretical analysis of the behavior-based control law, and the computational complexity as the swarm size grows. 
In addition, if robots can switch their roles to adapt to the task and environment, the robot swarm would be more autonomous. 

\section*{Appendix A\\ Analysis of Negotiation Protocols}

In Appendix A, we present the convergence analysis of the proposed negotiation protocols, including stationary protocols \eqref{Equ_posesti_leaderless} and \eqref{Equ_headesti_leaderless}, and time-varying protocols \eqref{Equ_posesti_leader} and \eqref{Equ_headesti_leader} in the main text. The analysis is detailed as follows.  

\subsection*{A.~Convergence of Stationary Negotiation}

We first discuss the nonsingularity of $\gamma_i$ as proofed below. 
\begin{proof}[Proof of Lemma~\ref{Thm_nonsigularity}]
	Recall that $\gamma_i=\sum\nolimits_{j\in\mathcal{N}_i}{a_{ij}}$. Since all the diagonal entries of the degree matrix is positive, $\gamma_i>0$ for all $i \in \mathcal{V}$. 
\end{proof}

Then, we present the convergence analysis of negotiation protocols \eqref{Equ_posesti_leaderless} and \eqref{Equ_headesti_leaderless} as follows. 

\begin{proof}[Proof of Theorem~1]
	Let $p_{\rm c}=[p_{{\rm c},1}^{\rm T},p_{{\rm c},2}^{\rm T},p_{{\rm c},3}^{\rm T},...,p_{{\rm c},n}^{\rm T}]^{\rm T},v_{\rm c}=[v_{{\rm c},1}^{\rm T},v_{{\rm c},2}^{\rm T},v_{{\rm c},3}^{\rm T},...,v_{{\rm c},n}^{\rm T}]^{\rm T} \in \mathbb{R}^{dn}$, where $n=n_{\rm robot}$. 
	The matrix-vector form of \eqref{Equ_posesti_leaderless} is 
	\begin{equation}
		\dot{v}_{\rm c}=-c_1\tilde{L}p_{\rm c}-c_2\tilde{L}v_{\rm c}-v_{\rm c}
		\label{Equ_posesti_leaderless_mtr}
	\end{equation}
	where $\tilde{L}=L\otimes I_d$. 
	Here, $I_d \in \mathbb{R}^{d\times d}$ is a $d$-dimensional identity matrix, and $L$ is the Laplacian of ${\left[a_{ij}\right]}_{n\times n}$. 
	Let $\bar p_{\rm c}=\frac{1}{n}\sum\nolimits_{i=1}^n{p_{{\rm c},i}}$, $\bar v_{\rm c}=\frac{1}{n}\sum\nolimits_{i=1}^n{v_{{\rm c},i}}$, and let $p_{\rm c}=\delta_{\rm p}+{\bf 1}_n\otimes\bar p_{\rm c}$, $v_{\rm c}=\delta_{\rm v}+{\bf 1}_n\otimes\bar v_{\rm c}$, where $\delta_{\rm p},\delta_{\rm v}\in \mathbb{R}^{dn}$ denote the negotiation errors and satisfy $\delta_{\rm v}=\dot{\delta}_{\rm p}$. 
	Since $a_{ij}=a_{ji}$ for all $i, j\in\mathcal{V}$, it can be obtained that $\dot{\bar v}_{\rm c}=\frac{1}{n}\sum\nolimits_{i=1}^n{\dot{v}_{{\rm c},i}}=-\bar v_{\rm c}$. 
	Substituting $p_{\rm c}$ and $v_{\rm c}$ into \eqref{Equ_posesti_leaderless_mtr} yields 
	\begin{align}
		\dot{\delta}_{\rm v}=-c_1\tilde{L}\delta_{\rm p}-(c_2\tilde{L}+I_{dn})\delta_{\rm v}. 
		\label{Equ_posesti_leaderless_err}
	\end{align}
	Consider the Lyapunov candidate $V=\frac{1}{2}(c_1\delta_{\rm p}^{\rm T}\tilde{L}\delta_{\rm p}+\delta_{\rm v}^{\rm T}\delta_{\rm v})$. 
	Differentiating $V$ with respect to $t$ and substituting \eqref{Equ_posesti_leaderless_err} into it give $\dot V=-\delta_{\rm v}^{\rm T}(c_2\tilde{L}+I_{dn})\delta_{\rm v}<0$ for $\delta_{\rm v}\neq 0$, which implies that $\dot V \rightarrow 0$ as $t\rightarrow \infty$. 
	It is further implied that $p_{{\rm c},i}(\infty)=\bar p_{\rm c}$ and $v_{{\rm c},i}(\infty)=0$. Similar to the proof of \eqref{Equ_posesti_leaderless}, $\varphi_{{\rm c},i}(\infty)=\frac{1}{n}\sum\nolimits_{i=1}^n{\varphi_{{\rm c},i}}$ and $\omega_{{\rm c},i}(\infty)=0$ hold. 
\end{proof}

\subsection*{B.~Convergence of Time-Varying Negotiation}

Next, we present the convergence analysis of negotiation protocols \eqref{Equ_posesti_leader} and \eqref{Equ_headesti_leader} as follows. 

\begin{proof}[Proof of Theorem~2]
	Define the new states $p_{{\rm c},n+1}\triangleq p_{\rm ref}$ and $v_{{\rm c},n+1}\triangleq v_{\rm ref}$. 
	Then, we rewrite the protocol \eqref{Equ_posesti_leader} as 
	\begin{align}
		\dot{v}_{{\rm c},i}=-\sum\limits_{j=1}^{n+1}{a_{ij}\Big(c_1 p_{{\rm c},ij}+c_2 v_{{\rm c},ij}-\dfrac{1}{\gamma_i}\dot{v}_{{\rm c},j}\Big)}
		\label{Equ_headesti_leader_var}
	\end{align} 
	where $a_{i,n+1}=1$ for informed robots and $a_{i,n+1}=0$ for uninformed ones. 
	In particular, $a_{n+1,n+1}=1$. Recall that $c_2=\mu_1+\frac{c_1}{\mu_1\gamma_i}$ in \eqref{Equ_posesti_leaderless}. 
	Reorganizing and multiplying $\gamma_i$ on both sides of \eqref{Equ_headesti_leader_var} yield 
	\begin{align*}
		\sum\limits_{j\in\mathcal{N}_i}{a_{ij}\Big(\frac{c_1}{\mu_1}v_{{\rm c},ij}+\dot{v}_{{\rm c},ij}\Big)}=-\alpha \sum\limits_{j\in\mathcal{N}_i}{a_{ij}\Big(\frac{c_1}{\mu_1}p_{{\rm c},ij}+v_{{\rm c},ij}\Big)}
	\end{align*}
	where $\alpha=\mu_1 \gamma_i>0$. 
	Let $e_i=\sum\nolimits_{j\in\mathcal{N}_i}{a_{ij}(\frac{c_1}{\mu_1}p_{{\rm c},ij}+v_{{\rm c},ij})}$ for $i$. 
	Then, we get $\dot e_i=-\alpha e_i$, which implies that $e_i\rightarrow 0$ as $t\rightarrow \infty$. 
	It is further implied that $p_{{\rm c},i}(\infty)=p_{\rm ref}$ and $v_{{\rm c},i}(\infty)=v_{\rm ref}$. 
	The convergence proof of protocol \eqref{Equ_headesti_leader} is similar to \eqref{Equ_posesti_leader}. 
	We have $\varphi_{{\rm c},i}(\infty)=\varphi_{\rm ref}$ and $\omega_{{\rm c},i}(\infty)=\omega_{\rm ref}$. 
\end{proof} 

\section*{Appendix B\\ Conversion of Acceleration Command}

Crazychoir provides an interface for converting the acceleration input $u_i$ into attitude signals that Crazyflie can receive. 
Denote $u_i$ as $u_i=[u_{x,i},u_{y,i},u_{z,i}]^{\rm T}$. 
Let $T_i$, $\varphi_i$, $\theta_i$, and $\phi_i$ represent the thrust, yaw angle, pitch angle, and roll angle of Crazyflie $i$, respectively. 
The conversion is given as follows: 
\begin{align*}
	\left\{ \begin{array}{l}
		T_i=n_z^{\rm T}\left(R(mu_i+mg n_z)\right) \vspace{0.2em}\\
		\varphi_i= \arctan \dfrac{-u_{x,i}u_{y,i}}{u_{y,i}^2+(u_{z,i}+g)^2} \vspace{0.2em}\\
		\theta_i = \arcsin \dfrac{-u_{x,i}(u_{z,i}+g)}{\|u_i+gn_z\|\sqrt{u_{y,i}^2+(u_{z,i}+g)^2}} \vspace{0.2em}\\
		\phi_i = \arctan \dfrac{-u_{y,i}\|u_i+gn_z\|}{(u_{z,i}+g)\sqrt{u_{y,i}^2+(u_{z,i}+g)^2}} 
	\end{array} \right.
\end{align*}
with rotation matrix $R$ from inertial frame to body frame 
\begin{align*}
	R=\left[\begin{matrix}
		1-2y^2-2z^2 & 2xy+2wz &  2xz-2wy\\
		2xy-2wz & 1-2x^2-2z^2 & 2yz+2wx\\
		2xz+2wy & 2yz-2wx & 1-2x^2-2y^2\\
	\end{matrix}\right]
\end{align*}
where $[x,y,z,w]^{\rm T}$ is quaternions and $n_z=[0,0,1]^{\rm T}$ is a unit vector. 
Here, $m$ is the mass of a Crazyflie and $g$ is the acceleration due to gravity. 
Note that the axis $z$ in both frames is pointing upward and forms a right-handed coordinate system with the axes $x$ and $y$. 

\section*{Appendix C\\ Discussion on Parameter Settings}

In Appendix C, we present more explanation of the parameter settings. 
The detailed discussion is as follows. 

\subsection*{A.~Parameter Classification}

The proposed strategy has a total of 11 parameters, including $r_{\rm body}$, $r_{\rm sense}$, $r_{\rm avoid}$, $c_1$, $c_3$, $\mu_1$, $\mu_2$, $\kappa_1$, $\kappa_2$, $\kappa_3$, and $\kappa_4$. 
According to their characteristics, these parameters can be divided into three classes. 

\begin{enumerate}
	\item Model parameters, including $r_{\rm body}$, $r_{\rm sense}$, and $r_{\rm avoid}$, are related to the robot model used. They are constrained by the robot's own capabilities, such as physical size and wireless transmission power. 
	\item Negotiation parameters, including $c_1$, $c_3$, $\mu_1$, $\mu_2$, determine the negotiation performance on shape position and orientation. 
	Here, $c_1, \mu_1$ are related to the position negotiation and $c_3, \mu_2$ determine the orientation one. 
	\item Control parameters, including $\kappa_1$, $\kappa_2$, $\kappa_3$, and $\kappa_4$, affect the ultimate performance of shape formation, which scale the shape-forming, shape-stabilizing, collision-avoidance, and velocity-alignment behaviors, respectively. 
\end{enumerate}

\vspace{-0.3em}
\subsection*{B.~Parameter Impact and Selection}

Model parameters are robot-fixed because they are completely constrained by the robot's capabilities. 
For example, if the robot's wireless transceiver only supports transmission within a range of $2.5{\rm m}$, then $r_{\rm sense}$ is chosen to be $2.5{\rm m}$. 
Different from the other two parameters, $r_{\rm avoid}$ requires manually setting and is between $r_{\rm body}$ and $r_{\rm sense}$. 
The parameter $r_{\rm avoid}$ affects the anti-collision behavior between adjacent robots. 
If $r_{\rm avoid}$ is too small, adjacent robots may collide due to the insufficient distance to react. 
If $r_{\rm avoid}$ is too large, robots may lose their neighbors due to the full effect of the anti-collision repulsion within the sensing range $r_{\rm sense}$. 
In experience, $r_{\rm avoid}$ is usually chosen to be about half of $r_{\rm sense}$. 

Negotiation parameters are insensitive in both simulation and experiment because the negotiation is a virtual process and free of real robot dynamics. 
In position negotiation, $c_1$ and $\mu_1$ scale the relative errors of position and velocity respectively, which affect the convergence process of negotiation. 
If $c_1$ is too large, the convergence process will be accelerated but encounter overshoot and oscillation. 
In contrast, if $\mu_1$ is too large, the convergence will be very smooth but take a very long time. 
Here, it is expected that the negotiation converges as quickly and smoothly as possible without overshoot. 
In experience, $\mu_1$ is selected to be approximately twice as large as $c_1$. 
To gain better performance, $c_1$ and $\mu_1$ need to be fine-tuned. 
The parameters $c_3$ and $\mu_2$ in orientation negotiation are selected to be the same as $c_1$ and $\mu_1$. 

Control parameters determine the shape control process and their settings are crucial to the performance. 
The scale of a parameter reflects the strength of the corresponding behavior. 
For example, if $\kappa_1$ becomes larger, the robot will move into the target shape faster, and otherwise, the robot will enter the shape slower. 
It is not always the case that the larger $\kappa_1$, the better. The reasons are as follows. 
First, a larger $\kappa_1$ causes oversaturation of the control input $u_i$ due to the limitations of real robot motion. 
Second, a larger $\kappa_1$ brings a large inertia to the robots. 
Thus, these robots may bring oscillations to other robots inside the shape while entering the target shape. 
Third, the setting of $\kappa_1$ is also affected by other control parameters. 
Note that the analysis of other parameters is similar to that of $\kappa_1$, so it will not be repeated here. 
All the control parameters are manually adjusted in the order of $\kappa_1$, $\kappa_3$, $\kappa_2$, and $\kappa_4$. 
More specifically, we first set other parameters to zero, and adjust $\kappa_1$ such that robots move into the shape as quickly while avoiding input saturation. 
Then, based on $\kappa_1$, we set $\kappa_3$ to ensure that all the robots are free of collisions. 
Based on $\kappa_1$ and $\kappa_3$, we next adjust $\kappa_2$ to stabilize all the robots in the target formation. 
Finally, we adjust $\kappa_4$ such that robots reach consistent motion as quickly as possible while forming the target shape. In experience, $\kappa_4$ is chosen to be $1$. 

\section*{Appendix D\\ Additional Simulation Figure}

In addition to the simulation figures shown in the main text, we present another figure in Appendix D, as shown in Fig.~\ref{Fig_velocityscalar}. 
As can be seen, each robot's velocity eventually becomes zero in all 5 simulation examples. 
One can conclude that all the robots remain stationary so as to save energy after forming the target shape in stationary formations. 

\begin{figure}[!h]
	\centering 
	\includegraphics[width=8.1cm]{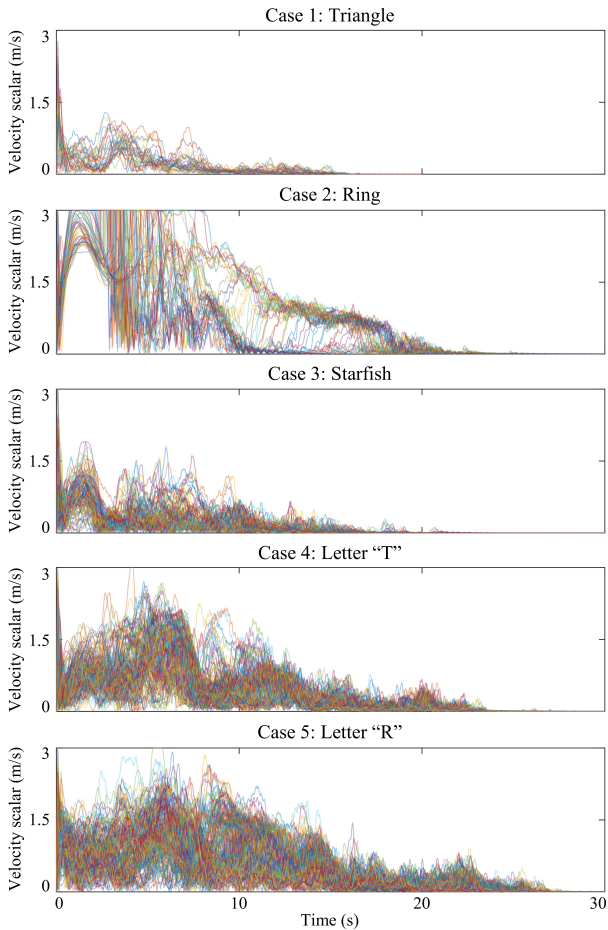}
	\caption{Velocity-scalar variations of 5 simulation examples in Fig.~3(a) in the main text. From top to bottom, the 5 diagrams correspond to cases 1 to 5, respectively.} 
	\label{Fig_velocityscalar}
\end{figure}


\bibliographystyle{bibliography/IEEEtranTIE}
\bibliography{bibliography/IEEEabrv,bibliography/references}\ 

\end{document}